\documentclass[conference]{IEEEtran}
\usepackage{cite}

\ifCLASSINFOpdf
   \usepackage[pdftex]{graphicx}
\else
\fi
\usepackage{array}

\usepackage{stfloats}
\usepackage{url}

\usepackage{algorithm,algpseudocode}

\usepackage{listings}
\usepackage{color}
\usepackage{subfig}
\definecolor{dkgreen}{rgb}{0,0.6,0}
\definecolor{gray}{rgb}{0.5,0.5,0.5}
\definecolor{mauve}{rgb}{0.58,0,0.82}

\renewcommand\IEEEkeywordsname{Keywords}

\begin{document}
%
\title{Neural Machine Translation System of Indic Languages - An Attention based Approach}

\author{\IEEEauthorblockN{Parth Shah}
\IEEEauthorblockA{
Uka Tarsadia University\\
Bardoli, India\\
parthpunita@yahoo.in}
\and
\IEEEauthorblockN{Vishvajit Bakrola}
\IEEEauthorblockA{
Uka Tarsadia University\\
Bardoli, India\\
vishvajit.bakrola@utu.ac.in}
}

%


\maketitle

\begin{abstract}
Neural machine translation (NMT) is a recent and effective technique which led to remarkable improvements in comparison of conventional machine translation techniques. Proposed neural machine translation model developed for the Gujarati language contains encoder-decoder with attention mechanism. In India, almost all the languages are originated from their ancestral language - Sanskrit. They are having inevitable similarities including lexical and named entity similarity. Translating into Indic languages is always be a challenging task. In this paper, we have presented the neural machine translation system (NMT) that can efficiently translate Indic languages like Hindi and Gujarati that together covers more than 58.49 percentage of total speakers in the country. We have compared the performance of our NMT model with automatic evaluation matrices such as BLEU, perplexity and TER matrix. The comparison of our network with Google translate is also presented where it outperformed with a margin of 6 BLEU score on English-Gujarati translation.

\end{abstract}


%
\IEEEpeerreviewmaketitle

\section{Introduction}

India is a highly diverse multilingual country in the world. In India, people of different regions use their own regional speaking languages, which makes India a country having world's second highest number of languages.  Human spoken languages in India belongs to several language families. Two main of those families are typically known as Indo-Aryan languages having 78.05 percentage Indian speakers \cite{census}  and Dravidian languages having 19.64 \cite{census} percentage Indian speakers. Hindi and Gujarati are among constitutional languages of India having nearly 601,688,479 \cite{census} Indian speakers almost 59 \cite{census} percentage of total country population. Constitute of India under Article 343 offers English as second additional official language having only 226,449 \cite{census} Indian speakers and nearly 0.02 percentages of total country population \cite{census}. Communication and information exchange among people is necessary for sharing knowledge, feelings, opinions, facts, and thoughts. Variation of English is used globally for human communication. The content available on the Internet is exceptionally dominated by English. Only 20 percent of the world population speaks in English, while in India it is only 0.02 \cite{census}. It is not possible to have a human translator in the country having this much language diversity. In order to bridge this vast language gap we need effective and accurate computational approaches, which require minimum human intervention. This task can be effectively done using machine translation.

Machine Translation (MT) is described as a task of computationally translate human spoken or natural language text or speech from one language to another with minimum human intervention. Machine translation aims to generate translations which have the same meaning as a source sentence and grammatically correct in the target language. Initial work on MT started in early 1950s \cite{sheridan1955research}, and has advanced rapidly since the 1990s due to the availability of more computational capacity and training data. Then after, number of approaches have been proposed to achieve more and more accurate machine translation as, Rule-based translation, Knowledge-based translation, Corpus-based translation, Hybrid translation, and Statistical machine translation(SMT) \cite{sheridan1955research}. All the approaches have their own merits and demerits. Among these, SMT which is a subcategory of Corpus based translation, is widely used as it is able to produce better results compared to other previously available techniques. The usage of the Neural networks in machine translation become popular in recent years around the globe and the novel technique of machine translation with the usage of neural network is known as Neural Machine Translation or NMT. In recent years, many works has been carried out on NMT. Little has been done on Indian languages as well \cite{sheridan1955research}. We found the NMT approach on Indic languages is still a challenging task, especially on bilingual machine translation.

In our past research work, we have worked on sequence-to-sequence model based machine translation system for Hindi language\cite{10.1007/978-981-10-8657-1_62}. In this work, we have improved that model and applied for English-Gujarati language pair. We have developed a system that uses neural model based on Attention mechanism. Our proposed attention based NMT model is tested with evaluation matrices as BLEU, perplexity and TER.

In section 2 overview of related work carried out in the domain of machine translation is described in brief, section 3 gives fundamentals of machine translation process with neural network using attention mechanism, section 4 gives a comparative analysis of various automatic evaluation matrices, section 5 introduce the proposed bilingual neural machine translation models, section 6 shows the implementation and generated results with our attention based NMT model is shown in section 7, conclusion of the paper is presented in section 8.



\section{Related work}
The process of translating text from source language to target language automatically with machine without any external human intervention is generally referred as Machine Translation(MT). It will basically convert sequence of words from source language to another sequence of words in target language without altering meaning of source words. Initial work in the field of machine translation was conceived by researchers at IBM research laboratory in the early '50s. They have also provided a successful demonstration in 1956 for machine translation system\cite{sheridan1955research}. But soon automatic language processing advisory committee of American government reported that machine translation task is infeasible to scale due to the amount of resource it requires. A new breakthrough in machine translation came only after 1979 where domain-specific translation system was implemented for weather bulletin translation from English to French\cite{Bostad_practicalexperience} \cite{Durand1991}.
In the year 1991, researchers from IIT Kanpur has developed Angla Bharati-I machine translation system \cite{sitender2012survey}\cite{dwivedi2010machine}. It was a general purpose translation system with domain customization. It is specifically designed for translating English to Hindi. In the year of 1999, CDAC developed a machine translation system named MANTRA \cite{sitender2012survey}, that uses the transfer-based machine translation. The system is developed for working on English-Gujarati, English-Hindi, English-Bengali and English-Telugu data pairs. Later the system is upgraded to AnglaBharati-II \cite{sitender2012survey}\cite{dwivedi2010machine} using a hybrid approach of machine translation in 2004. In AnglaBharati-II, the efficiency of the system is improved compared to AnglaBharati-I.

\section{Machine Translation}
Machine translation can be stated as the process of translating source language into target language considering the grammatical structure of the source language. The 1990s was marked as the breakthrough of a fairly new approaches to challenge and eventually improve the already established methodologies. This approach of machine translation was based on generating insights from large amount of available parallel corpuses. Example based Machine Translation was first proposed in 1981, but was developed from about 1990 onwards \cite{mitkov2005oxford}. The core idea is to reuse existing translations for generating a new translation\cite{somers2003machine}.

\subsection{Statistical Machine Translation}
Statistics based approach for machine translation does not utilize any traditional linguistic data. It basically works on the principle of probability. Here, the word in a source language corresponds to other similar word(s) in the given target language. However it requires a large corpus of reliable translations consisting in both source and target language sentences. This approach is similar to the methods of the IBM research group, which had initial success for speech recognition and Machine Translation in the early 1990s \cite{mitkov2005oxford}.

\subsection{Rule-based Machine Translation}
Normally all the languages used by humans for communication consist of certain amount of grammatical rules. If we are able to model these rules into a system, we can generate the natural fluent sentences in target language. Rule-based machine translation system tries to model the same approach for machine translation by mapping source and target language sentences using necessary rules. However to translate Indian languages large number of rules with different context are required \cite{indic}.

\subsection{Phrase-based Machine Translation}
A phrase is a small group of words which have some special meaning. Phrase-based machine translation system contains a phrase table, which has a list of translated sentences between source and target language. In addition to that, it is having information about how we can rearrange translation of multiple phrases to generate a meaningful target language sentence. But, these types of machine translation systems were unable to produce human-like natural language sentences as it is not possible to have all combination of different phrase every time in model\cite{indic}.

\subsection{Neural Machine Translation}\label{sec33}
\begin{figure}[h]
  \centering
  \includegraphics[width=9cm,height=6cm]{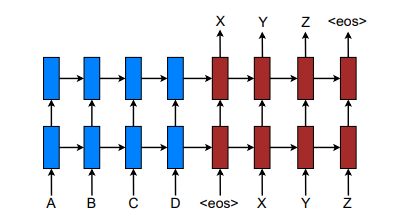}
  \caption{Converting source language into target language using sequence to sequence  model\cite{NIPS2014_5346}}\label{nmt_basic}
\end{figure}

Neural Machine Translation is one of the most recent approaches of machine translation that use a neural network based on the conditional probability of translating a given source language input to a given target language output as shown in Figure \ref{nmt_basic}. NMT is more appealing as it requires less knowledge related to the structure of source as well as target language. It has outperformed traditional MT models in large-scale translation tasks such as English to German and English to French \cite{DBLP:journals/corr/WuSCLNMKCGMKSJL16}. In recent years various architectures are proposed to achieve neural network based machine translation such as, simple encoder-decoder based model, RNN based model and LSTM model that learn problems with long-range temporal dependencies and the most advanced neural model for machine translation is Attention mechanism-based model.

Recurrent models typically factor computation along the symbol positions of the input and output sequences. Aligning the positions to steps in computation time, they generate a sequence of hidden states $h_t$, as a function of the previous hidden state $h_t+1$ and the input for position $t$ \cite{vaswani2017attention}.
This inherently sequential nature of RNN makes impossible to apply parallelization within training examples. But for longer sequence lengths, it becomes critical as memory constraints limits batching across examples\cite{DBLP:journals/corr/LuongPM15}. One of the major drawback of models that works on sequence-to-sequence model is that it is not able to generate words that are rarely encountered in input corpus. For solving this problem, attention mechanism can be applied in traditional sequence-to-sequence model.
It allows modeling of dependencies without regard to their distance in the input or output.
\begin{figure}
  \centering
  \includegraphics[width=9cm]{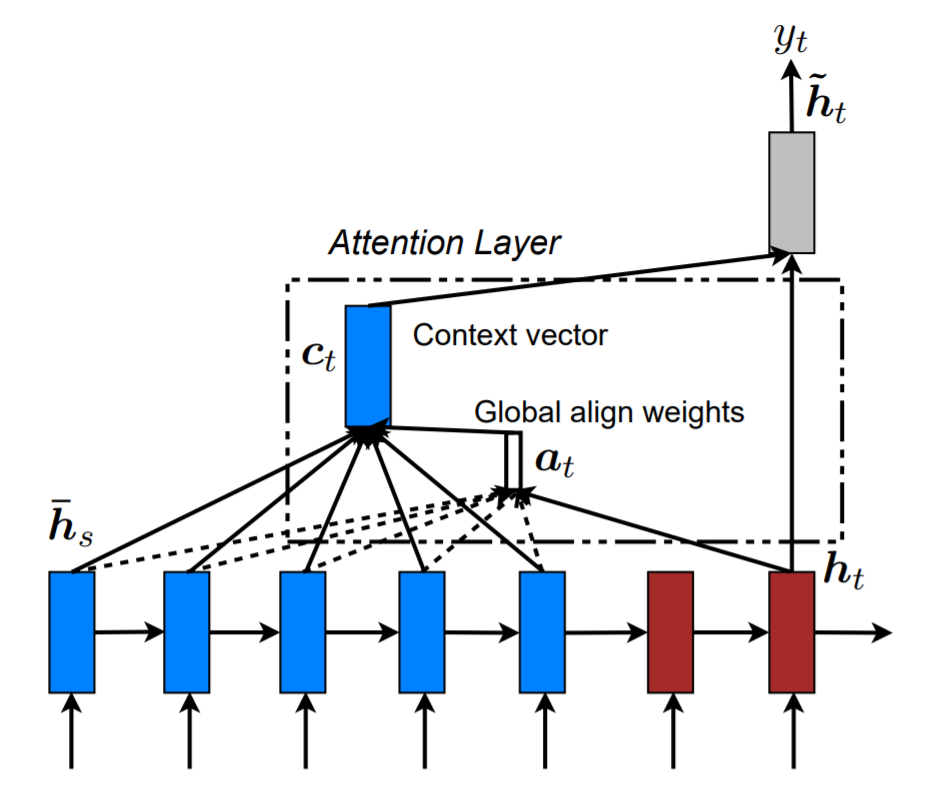}
  \caption{Basic structure of attention mechanism\cite{NIPS2014_5346}}\label{att}
\end{figure}
The concept of ``\textit{attention}" has gained popularity recently in training of neural networks, allowing models to learn alignments between different modalities, e.g., between image objects and agent actions in the dynamic control problem \cite{DBLP:journals/corr/LuongPM15}. As shown in Figure \ref{att}, it also provides context which will become helpful for generating more natural looking sentences including rare words. Recently, attentional NMT models have dominated the field of machine translation. They are pushing the boundary of translation performance by continuing new development in NMT architectures.

\section{Evaluation Matrices}
We can compare the performance of any machine translation model by comparing it across various evaluation matrices.
 In this paper, the following evaluation matrices are used for estimating the performance of our model.
\subsection{Translation error rate}
Translation error rate or TER measures the amount of editing it requires to match the human-generated output. It was designed for evaluating the output of machine translation avoiding the knowledge intensiveness of meaning-based approaches.  This method provides more meaningful insights when there is a large number of reference sentences available in the dataset. We can find TER of any translated sentences using the following equation \cite{snover2006study}:
\begin{equation}\label{bleu}
TER = \frac{Number\ of\ edits}{Average\ number\ of\ reference\ word}
\end{equation}

\subsection{Perplexity Matrix}
Perplexity is a measure of language model performance based on average probability. Perplexity can be defined as the inverse probability of the sentences available in test data, normalized by the number of words in generated sentences. It can be calculated using following equation \cite{jelinek1977perplexity}:
\begin{equation}\label{perp}
  PP_{T}(PM)=\frac{1}{\left(\prod_{i=1}^{t}PM\left(w_{i}\mid w_{1}\cdots w_{i-1}\right)\right)^{\frac{1}{2}}}
\end{equation}

\subsection{BLEU}
BLEU uses the basic concepts of n-gram precision to calculate similarity between reference and generated sentence. It correlates highly with human expert review as it uses the average score of all result in test dataset rather than providing result of each sentence. BLEU score can be computed using the following equation \cite{papineni2002bleu}:
\begin{equation}\label{bleu}
  p_n=\frac{\sum_{C\in\{Candidates\}}\sum_{ngram\in C}Count_{clip}(ngram)}{\sum_{C'\in\{Candidates\}}\sum_{ngram'\in C'}Count(ngram')}
\end{equation}

\section{Proposed System}

\begin{figure}[h]
  \centering
  \includegraphics[width=9cm,height=6cm]{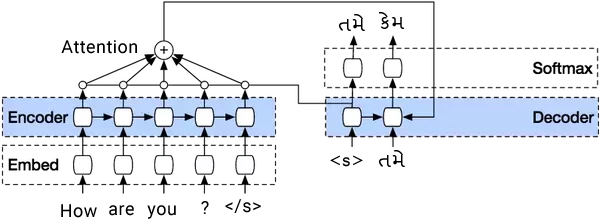}
  \caption{Proposed system using Attention Mechanism}\label{seq2seqwithatt}
\end{figure}

As shown in Figure \ref{seq2seqwithatt}, our proposed model is divided into mainly three different parts. Encoder, Decoder and Attention mechanism. Our encoder has two LSTM layers with 128 units of LSTM cells. This encoder will output encoded word embedding vector. This embedding vector is provided as input to decoder. Decoder is also consist of two LSTM layers with 128 units of lstm cells. It will take encoded vector and produce the output using beam search method. Whenever any output is produced the value of hidden state is compared with all input states to derive weights for attention mechanism. Based on attention weights, context vector is calculated and it is given as additional input to decoder for generating context relevant translation based on previous outcomes.

\section{Implementation}

\subsection{Datasets}
In order to  work with neural networks we require large amount of training data. As neural networks are learning with experience, more the experience accurate the learning is. Wide range of work has been carried out for non Indian languages. So enough amount of parallel corpus is available like English-French, English German, etc. But on Indian languages most of corpus was available only for English-Hindi language pair. The only dataset available for Gujarati language is OPUS\cite{TIEDEMANN12.463}, which is a collection of translated texts from user manual of the open source software. So in order to create machine translation system that works on conversational level we have created our new dataset. The created \textit{"eng\_guj\_parallel\_corpus"} contains nearly 65000 sentences in parallel format. We have also made it available for all researchers as open source dataset and can be downloaded from  \url{https://github.com/shahparth123/eng_guj_parallel_corpus}. It is collection of sentences describing the activity or scenes in both Gujarati and English language.

\subsection{Experiment Setup}
For our experiment we have used Google Cloud's n1-highmem-2 instance with Intel Xeon E5 processor, 13 GB of primary memory and Tesla K80(2496 CUDA Core) GPU with 12GB of GPU memory. For creating and training deep neural networks TensorFlow deep learning library is used\cite{abadi2016tensorflow}.

\section{Results and Discussion}
\subsection{Results}

In our experiment we have trained our proposed neural machine translation model using \textit{"eng\_guj\_parallel\_corpus"} with 37000 epoch. Some of the results for proposed model is given in following Figure \ref{figop1} and \ref{figo} :
\begin{figure}[h]
  \centering
  \includegraphics[width=9cm]{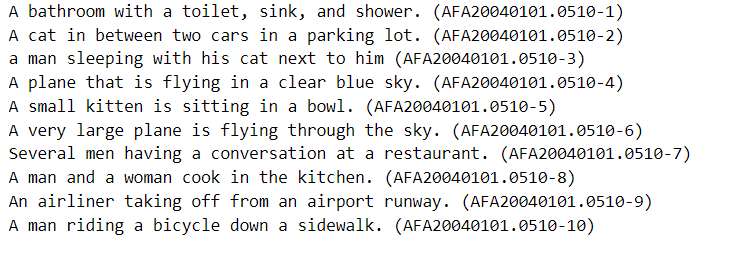}
  \caption{Input data}\label{figop1}
\end{figure}

\begin{figure}[h]
  \centering
  \includegraphics[width=9cm]{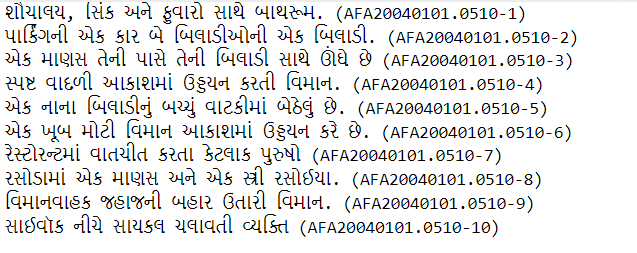}
  \caption{Generated output data}\label{figo}

\end{figure}
As seen in figures, in most of the cases our model produces comparable result with human translator.
Result for BLEU score for our model and Google's Neural Machine Translation is compared in table \ref{tabop}:
\begin{table}[H]
\caption{Various evaluation matrix comparison of models\label{tabop}}
\centering{}%
\begin{tabular}{|>{\centering}m{2.5cm}|>{\centering}m{3cm}|>{\centering}m{2cm}|}
  \hline
  Evaluation Matrix  & Proposed Model & GNMT \tabularnewline
  \hline
  BLEU & 40.33  & 33.66 \tabularnewline
  \hline
  TER & 0.3913 & 0.5217 \tabularnewline
  \hline
  Perplexity & 2.37 & - \tabularnewline
  \hline
\end{tabular}
\end{table}

\section{Conclusion}
The conventional machine translation approaches are fast and efficient enough in processing. They have been proven significant in delivering good accuracy with their limited scope of application. But, they are facing difficulties in generating a target sentence or corpus with human-like fluency. Neural machine translation has played a significant role to outperformed difficulties associated with conventional machine translation approaches. However, the NMT models widely used in recent years like Seq-2-Seq has given great accuracy in generating fluent target language. Though, on some real environmental situations, specifically in the case when the model is coming across rare words the accuracy is decreased dramatically. To overcome this limitation of recent NMT models, an Attention mechanism is equipped in the model that has improved the accuracy. We have achieved an average BLEU score of 59.73 on training corpus and 40.33 on test corpus from parallel English-Gujarati corpus having 65,000 sentences.

%
%
%
%

\bibliographystyle{IEEEtran}
\bibliography{reference}
%

\end{document}